\documentclass[runningheads]{llncs}
\usepackage[utf8]{inputenc} % allow utf-8 input
\usepackage[T1]{fontenc}    % use 8-bit T1 fonts
\usepackage{hyperref}       % hyperlinks
\usepackage{url}            % simple URL typesetting
\usepackage{booktabs}       % professional-quality tables
\usepackage{amsfonts}       % blackboard math symbols
\usepackage{nicefrac}       % compact symbols for 1/2, etc.
\usepackage{microtype}      % microtypography
\usepackage{lipsum}		% Can be removed after putting your text content
\usepackage{graphicx}
\usepackage{tabularx}
\usepackage{doi}
\usepackage{subcaption}

\usepackage{threeparttable}
\begin{document}

\title{Combining EfficientNet and Vision Transformers for Video Deepfake Detection}
\titlerunning{Combining EfficientNet and ViTs for Video Deepfake Detection}
% If the paper title is too long for the running head, you can set
% an abbreviated paper title here
%
%\author{Anonymous Submission}
\author{Davide Coccomini\ \and
Nicola Messina \and
Claudio Gennaro \and Fabrizio Falchi}

\authorrunning{D. Coccomini et al.}
% First names are abbreviated in the running head.
% If there are more than two authors, 'et al.' is used.
%
\institute{
ISTI-CNR, 
via G. Moruzzi 1, 56124,
Pisa, Italy
\email{davidealessandro.coccomini@isti.cnr.it, nicola.messina@isti.cnr.it, claudio.gennaro@isti.cnr.it, @fabrizio.falchi@isti.cnr.it}}

\maketitle              % typeset the header of the contribution
\begin{abstract}
	Deepfakes are the result of digital manipulation to forge realistic yet fake imagery. With the astonishing advances in deep generative models, fake images or videos are nowadays obtained using variational autoencoders (VAEs) or Generative Adversarial Networks (GANs). These technologies are becoming more accessible and accurate, resulting in fake videos that are very difficult to be detected. Traditionally, Convolutional Neural Networks (CNNs) have been used to perform video deepfake detection, with the best results obtained using methods based on EfficientNet B7. In this study, we focus on video deep fake detection on faces, given that most methods are becoming extremely accurate in the generation of realistic human faces. Specifically, we combine various types of Vision Transformers with a convolutional EfficientNet B0 used as a feature extractor, obtaining comparable results with some very recent methods that use Vision Transformers. Differently from the state-of-the-art approaches, we use neither distillation nor ensemble methods. Furthermore, we present a straightforward inference procedure based on a simple voting scheme for handling multiple faces in the same video shot. The best model achieved an AUC of 0.951 and an F1 score of 88.0\%, very close to the state-of-the-art on the DeepFake Detection Challenge (DFDC). The code for reproducing our results is publicly available here: \url{https://github.com/davide-coccomini/Combining-EfficientNet-and-Vision-Transformers-for-Video-Deepfake-Detection}.
\end{abstract}

% keywords can be removed

\section{Introduction}
With the recent advances in generative deep learning techniques, it is nowadays possible to forge highly-realistic and credible misleading videos. These methods have generated numerous fake news or revenge porn videos, becoming a severe problem in modern society \cite{looming}. 
These fake videos are known as \textit{deepfakes}.
Given the astonishing realism obtained by recent models in the generation of human faces, deepfakes are mainly obtained by transposing one person's face onto another's. The results are so realistic that it is almost like the person being replaced is actually present in the video, and the replaced actors are rigged to say things they never actually said \cite{tolosana2020deepfakes}.

The evolution of deepfakes generation techniques and their increasing accessibility forces the research community to find effective methods to distinguish a manipulated video from a real one. At the same time, more and more models based on Transformers are gaining ground in the field of Computer Vision, showing excellent results in image processing \cite{khan2021transformers,han2020survey}, document retrieval \cite{macavaney2020efficient}, and efficient visual-textual matching  \cite{messina2020fine,messina2021transformer}, mainly for use in large-scale multi-modal retrieval systems \cite{amato2021visione,messina2021towards}.

In this paper, we analyze different solutions based on combinations of convolutional networks, particularly the EfficientNet B0, with different types of Vision Transformers and compare the results with the current state-of-the-art. Unlike Vision Transformers, CNNs still maintain an important architectural prior, the spatial locality, which is very important for discovering image patch abnormalities and maintaining good data efficiency. CNNs, in fact, have a long-established success on many tasks, ranging from image classification \cite{foret2020sharpness,xie2017aggregated} and object detection \cite{redmon2018yolov3,amato2019learning,ciampi2020virtual} to abstract visual reasoning \cite{messina2019testing,messina2021solving}.

In this paper, we also propose a simple yet effective voting mechanism to perform inference on videos. We show that this methodology could lead to better and more stable results.

\section{Related Works}
\subsection{Deepfake Generation}
There are mainly two generative approaches to obtain realistic faces: Generative Adversarial Networks (GANs) \cite{goodfellow2014generative} and Variational AutoEncoders (VAEs) \cite{kingma2014autoencoding}. 

GANs employ two distinct networks. The discriminator, the one that must be able to identify when a video is fake or not, and the generator, the network that actually modifies the video in a sufficiently credible way to deceive its counterpart.
With GANs, very credible and realistic results have been obtained, and over time, numerous approaches have been introduced such as StarGAN \cite{choi2018stargan} and DiscoGAN \cite{kim2017learning}; the best results in this field have been obtained with StyleGAN-V2 \cite{karras2020analyzing}.

VAE-based solutions, instead, make use of a system consisting of two encoder-decoder pairs, each of which is trained to deconstruct and reconstruct one of the two faces to be exchanged. Subsequently, the decoding part is switched, and this allows the reconstruction of the target person's face. %, thus obtaining the desired result. 
The best-known uses of this technique were DeepFaceLab \cite{deepfacelab}, DFaker\footnote{\url{https://github.com/dfaker/df}}, and DeepFaketf\footnote{\url{https://github.com/StromWine/DeepFake_tf}}.

\subsection{Deepfake Detection}
The problem of deepfake detection has a widespread interest not only in the visual domain. For example, the recent work in \cite{fagni2021tweepfake} analyzes deepfakes in tweets for finding and defeating false content in social networks.

In an attempt to address the problem of deepfakes detection in videos, numerous datasets have been produced over the years. These datasets are grouped into three generations, the first generation consisting of DF-TIMIT \cite{korshunov2018deepfakes}, UADFC \cite{yang2019exposing} and FaceForensics++ \cite{rossler2019faceforensics}, the second generation datasets such as Google Deepfake Detection Dataset \cite{googledf}, Celeb-DF \cite{li2020celebdf}, and finally the third generation datasets, with the DFDC dataset \cite{dolhansky2020deepfake} and DeepForensics \cite{jiang2020deeperforensics10}. The further the generations go, the larger these datasets are, and the more frames they contain.

In particular, on the DFDC dataset, which is the largest and most complete, multiple experiments were carried out trying to obtain an effective method for deepfake detection. Very good results were obtained with EfficientNet B7 ensemble technique in 
\cite{dfdc_solution}. Other noteworthy methods include those conducted in \cite{montserrat2020deepfakes}, who attempted to identify spatio-temporal anomalies by combining an EfficientNet with a Gated Recurrent Unit (GRU). Some efforts to capture spatio-temporal inconsistencies were made in  \cite{delima2020deepfake} using 3DCNN networks and in \cite{amerini2019deepfake}, which presented a method that exploits optical flow to detect video glitches. Some more classical methods have also been proposed to perform deepfake detection. In particular, the authors in \cite{guarnera2020deepfake} proposed a method based on K-nearest neighbors, while the work in \cite{yang2019exposing} exploited SVMs. Of note is the very recent work of Giudice et al. \cite{jimaging7080128} in which they presented an innovative method for identifying so-called GAN Specific Frequencies (GSF) that represent a unique fingerprint of different generative architectures. By exploiting the Discrete Cosine Transform (DCT) they manage to identify anomalous frequencies.

More recently, methods based on Vision Transformers have been proposed. Notably, the method presented in \cite{wodajo2021deepfake} obtained good results by combining Transformers with a convolutional network, used to extract patches from faces detected in videos. 

State of the art was then recently improved by performing distillation from the EfficientNet B7 pre-trained on the DFDC dataset to a Vision Transformer \cite{heo2021deepfake}. In this case, the Vision Transformer patches are combined with patches extracted from the EfficientNet B7 pre-trained via global pooling and then passed to the Transformer Encoder. A distillation token is then added to the Transformer network to transfer the knowledge acquired by the EfficientNet B7. % was added that is trained by the teacher network, EfficientNet B7.

\section{Method}
The proposed methods analyze the faces extracted from the source video to determine whenever they have been manipulated. For this reason, faces are pre-extracted using a state-of-the-art face detector, MTCNN \cite{zhang2016joint}. We propose two mixed convolutional-transformer architectures that take as input a pre-extracted face and output the probability that the face has been manipulated. The two presented architectures are trained in a supervised way to discern real from fake examples. For this reason, we solve the detection task by framing it as a binary classification problem. Specifically, we propose the \textit{Efficient ViT} and the \textit{Convolutional Cross ViT}, better explained in the following paragraphs.

The proposed models are trained on a face basis, and then they are used at inference time to draw a conclusion on the whole video shot by aggregating the inferred output both in time and across multiple faces, as explained in Section \ref{sec:inference}.

\begin{figure}[t]
\begin{subfigure}[b]{0.38\textwidth}
\centering
\includegraphics[width=\linewidth]{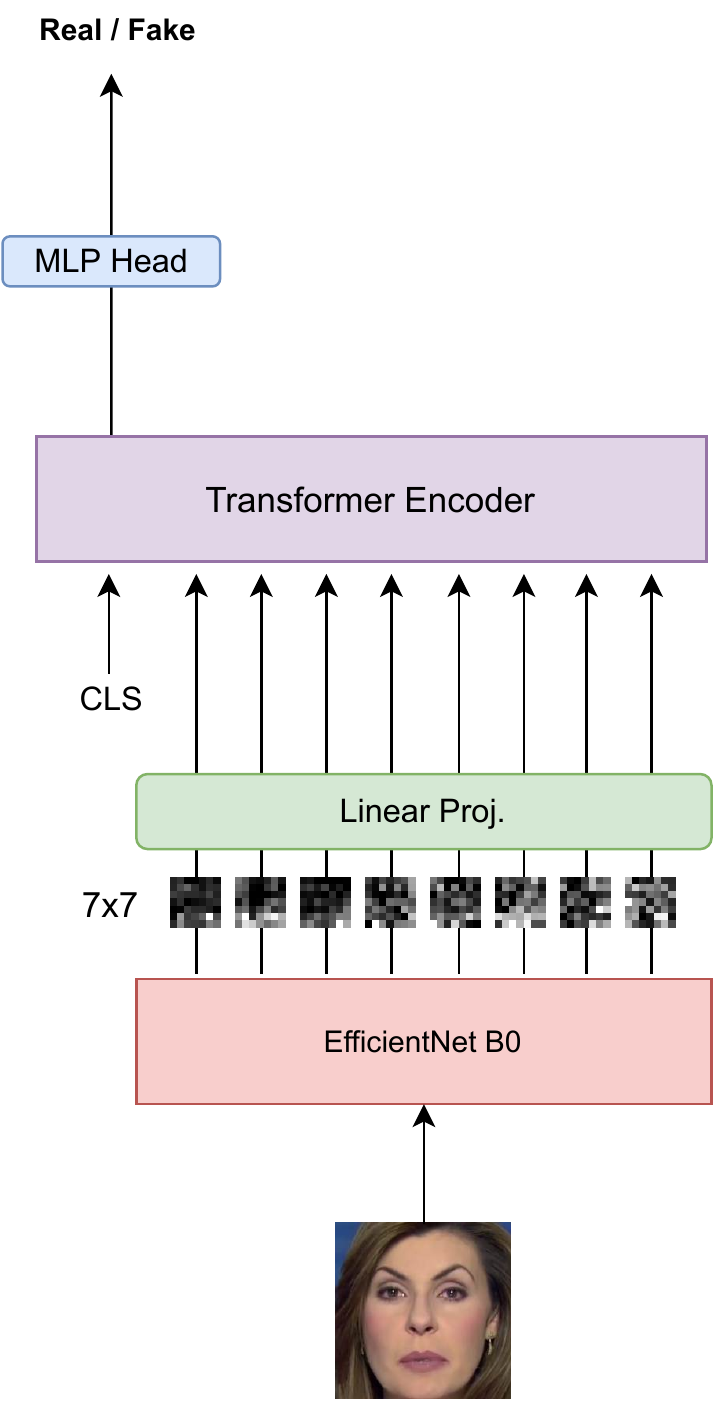}
\caption{Efficient ViT architecture.}
    \label{figure:vit}
\end{subfigure}
\hspace{0.5cm}
\begin{subfigure}[b]{0.5706\textwidth}
\centering
\includegraphics[width=\linewidth]{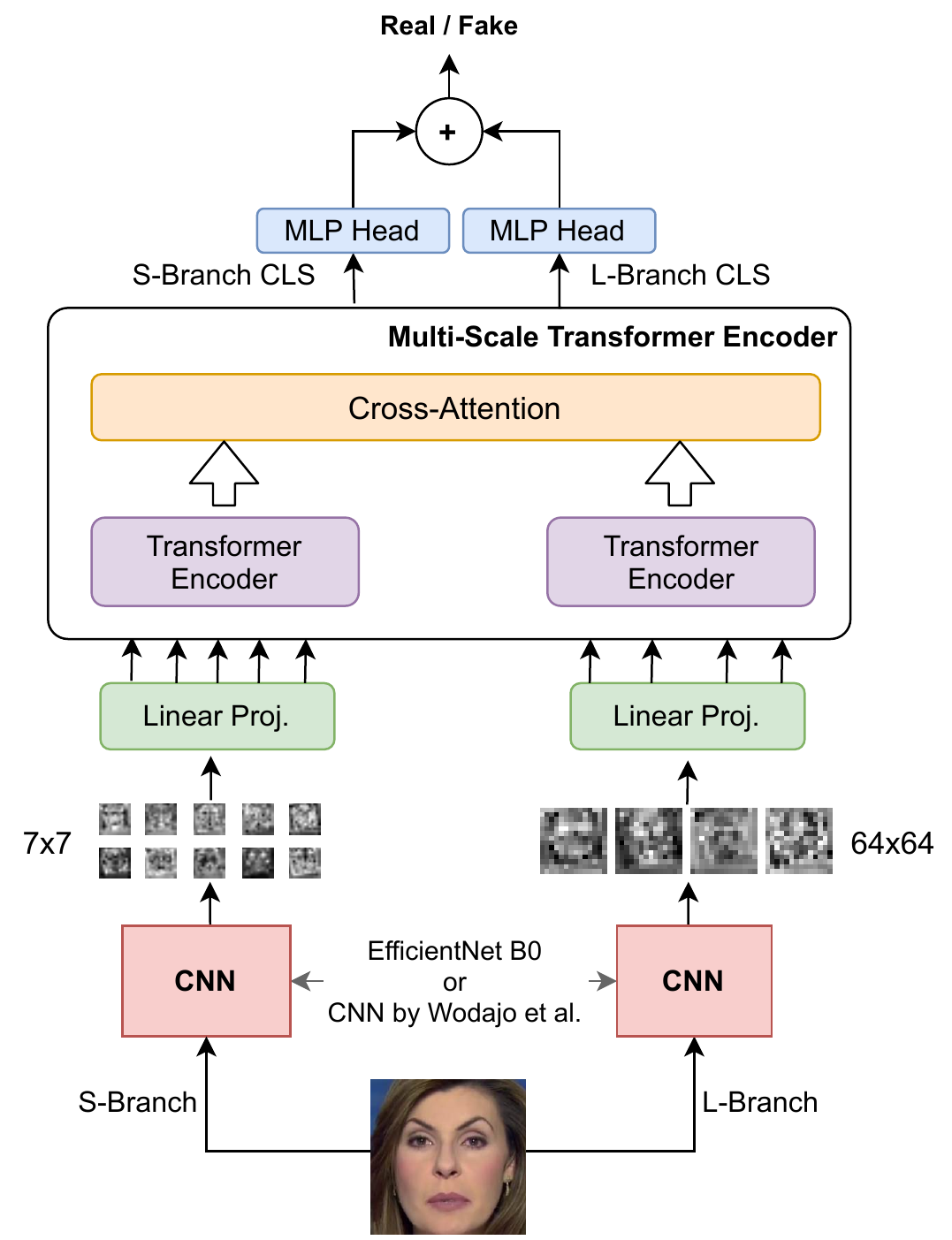}
\caption{Convolutional Cross ViT architecture.}
\label{figure:cross_vit}
\end{subfigure}
\caption{The proposed architectures. Notice that for the Convolutional Cross ViT in (b), we experimented both with EfficientNet B0 and with the convolutional architecture by \cite{wodajo2021deepfake} as feature extractors.}
%\label{fig:overfit}       % Give a unique label
\end{figure}

\paragraph{The Efficient ViT}
The Efficient ViT is composed of two blocks, a convolutional module for working as a feature extractor and a Transformer Encoder, in a setup very similar to the Vision Transformer (ViT) \cite{dosovitskiy2020image}.
Considering the promising results of the EfficientNet, we use an EfficientNet B0, the smallest of the EfficientNet networks, as a convolutional extractor for processing the input faces. Specifically, the EfficientNet produces a visual feature for each chunk from the input face. Each chunk is $7\times7$ pixels. After a linear projection, every feature from each spatial location is further processed by a Vision Transformer. The CLS token is used for producing the binary classification score. %We call this architecture Efficient ViT, 
The architecture is illustrated in Figure \ref{figure:vit}. %The faces extracted via MTCNN from a video deepfake are transformed by the EfficientNet into 7x7 representations, then sent to the Vision Transformer after a linear projection. %used within the self-attention mechanism of the Vision Transformer following linear projection.
% \begin{figure}[t]
%     \centering
%     \includegraphics[width=0.65\textwidth]{images/efficientvit.png}
%     \caption{EfficientNet B0 combined with Vision Transformer Architecture.}
%     \label{figure:efficient_vit}
% \end{figure}
The EfficientNet B0 feature extractor is initialized with the pre-trained weights and fine-tuned to allow the last layers of the network to perform a more consistent and suitable extraction for this specific downstream task. The features extracted from the EfficientNet B0 convolutional network simplify the training of the Vision Transformer, as the CNN features already embed important low-level and localized information from the image. %and they allow it to work on local-aware descriptions representations % that already make sense from the first epoch.

\paragraph{The Convolutional Cross ViT}
Limiting the architecture to the use only small patches as in the Efficient ViT may not be the ideal choice, as artifacts introduced by deepfakes generation methods may arise both locally and globally. 
For this reason, we also introduce the Convolutional Cross ViT architecture.
%Therefore, the structure above is further evolved using the same patch extractor but with a different type of Vision Transformer, the Cross Vision Transformer. 
The Convolutional Cross ViT builds upon both the Efficient ViT and the multi-scale Transformer architecture by \cite{chen2021crossvit}. More in detail, the Convolutional Cross ViT uses two distinct branches: the \textit{S-branch}, which deals with smaller patches, and the \textit{L-branch}, which works on larger patches for having a wider receptive field. The visual tokens output by the Transformer Encoders from the two branches are combined through cross attention, allowing direct interaction between the two paths. Finally, the CLS tokens corresponding to the outputs from the two branches are used to produce two separate logits. These logits are summed, and a final sigmoid produces the final probabilities. %Differently from the work by \cite{chen2021crossvit}, this architecture extracts image patches using the EfficientNet B0 network. 
A detailed overview of this architecture is shown in Fig. \ref{figure:cross_vit}.%to influence each other. 
%
% \begin{figure}[t]
%     \centering
%     \includegraphics[width=0.35\textwidth]{images/crossefficient.png}
%     \caption{EfficientNet B0 combined with Cross Vision Transformer Architecture.}
%     \label{figure:cross_efficient_vit}
% \end{figure}
For the Convolutional Cross ViT, we use two different CNN backbones. The former is the EfficientNet B0, which processes $7\times7$ image patches for the S-branch and $54\times54$ for the L-branch. The latter is the CNN by Wodajo et al. \cite{wodajo2021deepfake}, which handles $7\times7$ image patches for the S-branch and $64\times64$ for the L-branch.

\section{Experiments}

We probed the presented architectures against some state-of-the-art methods on two widely-used datasets. In particular, we considered Convolutional ViT \cite{wodajo2021deepfake}, ViT with distillation \cite{heo2021deepfake}, and Selim EfficientNet B7 \cite{dfdc_solution}, the winner of the Deep Fake Detection Challenge (DFDC).
Notice that the results for Convolutional ViT \cite{wodajo2021deepfake} are not reported in the original paper, but they are obtained executing the test code on DFDC test set using the available pre-trained model released by the authors.

\subsection{Datasets and Face Extraction}

%We conducted the tests on FaceForensics++ and on the 5000 test videos made available %for the DFDC dataset. In order to compare our methods also on the DFDC test set, we %tested the Convolutional Vision Transformer \cite{wodajo2021deepfake} on these videos %obtaining the necessary AUC and F1-score values for comparison. 

We initially conducted some tests on FaceForensics++. The dataset is composed of original and fake videos generated through different deepfake generation techniques. For evaluating, we considered the videos generated in the Deepfakes, Face2Face, FaceShifter, FaceSwap and NeuralTextures sub-datasets. We also used the DFDC test set containing 5000 videos. The model trained on the entire training set, which includes fake videos of all considered methods of FaceForensics++ and the training videos of DFDC dataset, was used to calculate the accuracy measures of the model, reported separately. In order to compare our methods also on the DFDC test set, we tested the Convolutional Vision Transformer \cite{wodajo2021deepfake} on these videos obtaining the necessary AUC and F1-score values for comparison.

% For training faces were extracted from the videos using an MTCNN and data augmentation inspired by that carried out in Selim \cite{dfdc_solution} was applied but with images extracted so that they were always square and without padding. 
During training, we extracted the faces from the videos using an MTCNN, and we performed data augmentation like in \cite{dfdc_solution}. Differently from them, we extracted the faces so that they were always squared and without padding. The images obtained are used during the training, thus ignoring the remaining part of the frames.
We used the Albumentations library \cite{2018arXiv180906839B}, and we applied common transformations such as the introduction of blur, Gaussian noise, transposition, rotation, and various isotropic resizes during training. 

\subsection{Training}

We trained the networks on 220,444 faces extracted from the videos of DFDC training set and FaceForensics++ training videos, and we used 8070 faces for validation from DFDC dataset. The training set was constructed trying to maintain a good balance between the real class composed of 116,950 images and fakes with 103,494 images. 

We used pre-trained EfficientNet B0 and Wodajo CNN feature extractors. However, we observed better results when fine-tuning them, so we did not freeze the extraction layers.
We used the standard binary cross-entropy loss as our objective during training. We optimized our network end-to-end, using an SGD optimizer with a learning rate of 0.01.

% For the testing of all our models instead we opted for a slightly more peculiar approach that sets a threshold for deciding whether or not a video is fake at 0.55 as done in Heo et al. \cite{heo2021deepfake} but instead of averaging all ratings on individual faces indistinctly within the video, it obtains summary values of the individual actors that are identified. If a group of faces is fake, then the whole video is fake as shown in Fig. \ref{figure:evaluation_strategy}.

\subsection{Inference}
\label{sec:inference}
\begin{figure}[t]
\begin{subfigure}[b]{0.54\textwidth}
\centering
\includegraphics[width=\textwidth]{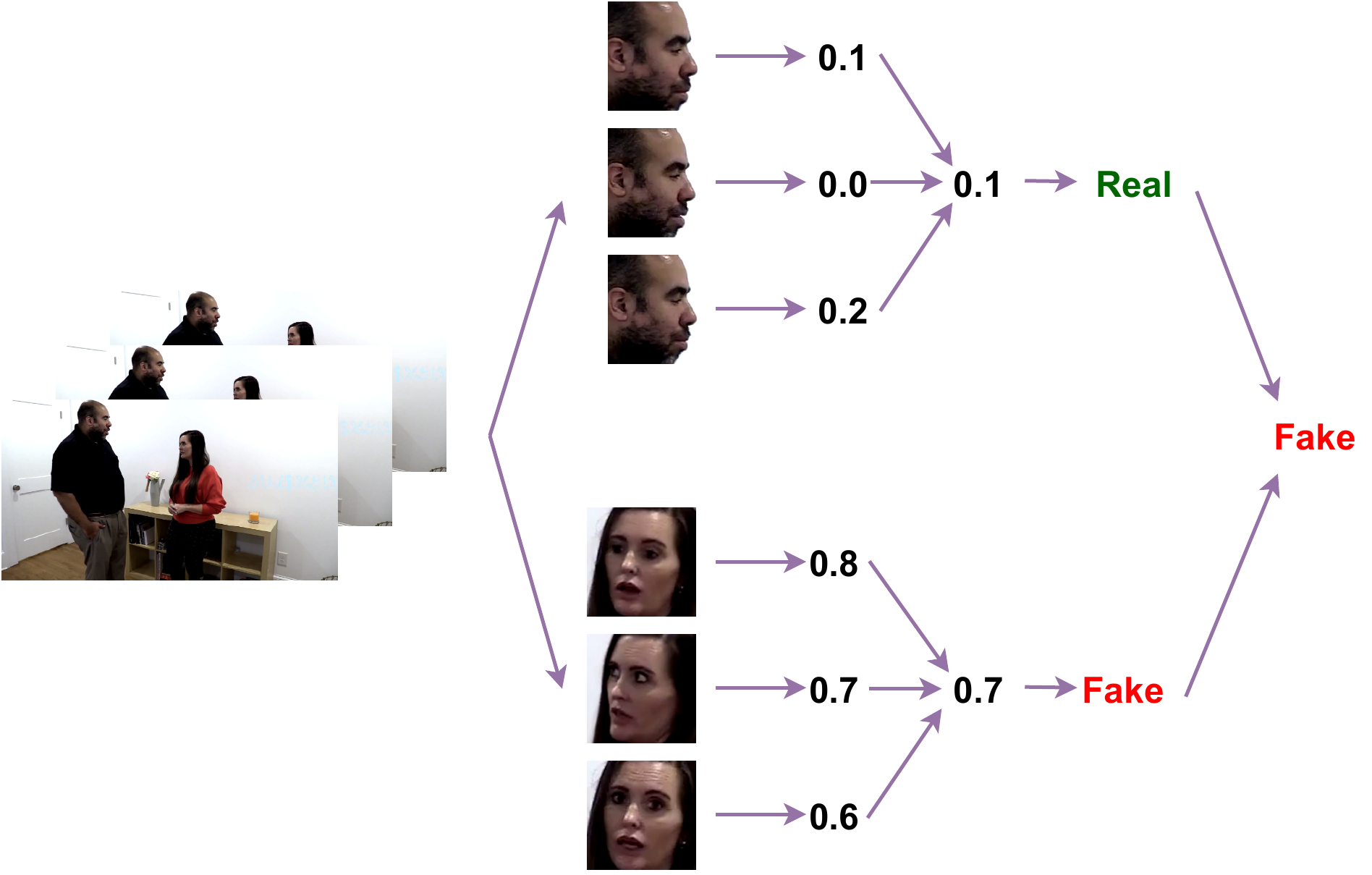}
\caption{Inference strategy with multiple faces in the same video.}
\label{figure:evaluation_strategy}
\end{subfigure}
\begin{subfigure}[b]{0.46\textwidth}
\centering
\includegraphics[width=\textwidth]{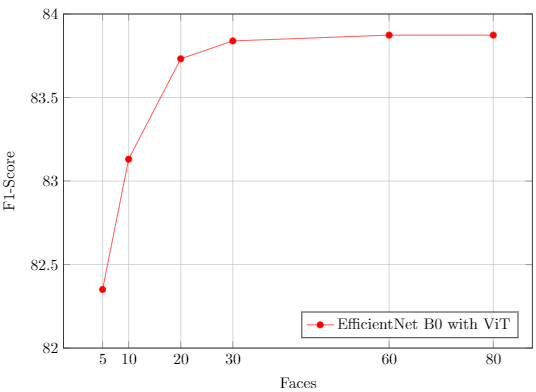}
\caption{F1-score versus the number of extracted faces.}
\label{figure:faces_choice}
\end{subfigure}
\caption{Inference.}
%\label{fig:overfit}       % Give a unique label
\end{figure}

At inference time, we set a real/fake threshold at 0.55 as done in \cite{heo2021deepfake}. However, we proposed a slightly more elaborated voting procedure instead of averaging all ratings on individual faces indistinctly within the video. Specifically, we merged the scores, grouping them by the identifier of the actors. The face identifier is available as an output from the employed MTCNN face detector. The scores from different actors are averaged over time to produce a probability of the face being fake. Then, the per-actor scores are merged using hard voting. In particular, if there is at least one actor face passing the threshold, the whole video is classified as fake. The procedure is graphically explained in Fig. \ref{figure:evaluation_strategy}. 
We claim that this approach is helpful to handle better videos in which only one of the actors' faces has been manipulated.% still looks real due to the threshold drop caused by the other actors in the scene.
%This peculiarity in the evaluation could also be exploited by an attacker by manipulating videos in which many people are present and changing only one face. 
%Without considering faces separately, the entire video would most likely be considered real, bypassing deepfake detection systems.

% \begin{figure}[t]
%     \centering
%     \includegraphics[width=0.6\textwidth]{images/inference.pdf}
%     \caption{Inference strategy.}
%     \label{figure:evaluation_strategy}
% \end{figure}

%During face extraction, it is possible that the MTCNN will not return any results due to the presence within the test set of videos with poor lighting or containing distractors that make face detection more difficult. In cases where , we gradually reduce the face detection threshold by accepting  %in the videos where it is difficult to obtain a result, 
%even the faces with lower detection confidence scores. 

In addition, it is interesting to evaluate how the performance changes when a varying number of faces are considered at inference time. To ensure that the tests are as light yet effective as possible, we experimented on one of our networks to see how the F1-score varies with the number of faces considered at testing time (Fig. \ref{figure:faces_choice}). We notice that a plateau is reached when no more than 30 faces are used, so employing more than this number of faces seems statistically useless at inference time.
%
% \begin{figure}[t]
%     \centering
%     \includegraphics[width=0.4\textwidth]{images/faces_choice.png}
%     \caption{f1-score variation on considered faces}
%     \label{figure:faces_choice}
% \end{figure}
%
%As can be seen from the plot in Fig. \ref{figure:faces_choice} the maximum F1-score is achieved using 30 faces. After this threshold, the F1-score remains almost unchanged. %even for a larger number of faces.% such as 80.

\subsection{Results}
Table \ref{tab:dfdc_results} shows that all models developed with EfficientNet achieve considerably higher AUC and F1-scores than the Convolutional ViT presented in  \cite{wodajo2021deepfake}, providing initial evidence that this specific network structure may be more suitable for this type of task.
It can also be noticed that the models based on Cross Vision Transformer obtain the best results, confirming the theory that joined local and global image processing brings to better anomaly identification. %it is winning to work simultaneously both locally and globally. 

\newcolumntype{C}{>{\centering\arraybackslash}p{1.8cm}}
%\newcolumntype{L}{>{\raggedright\arraybackslash}p{3.2cm}}
\begin{table}[t]
\centering
\caption{Results on DFDC test dataset}
\label{tab:dfdc_results}
\begin{threeparttable}
\begin{tabular}{lCCC}
    \toprule
    \textbf{Model}    & \textbf{AUC} & \textbf{F1-score} & \textbf{\# params} \\
    \midrule
    ViT with distillation \cite{heo2021deepfake}      & 0.978  & 91.9\% & 373M\\
    Selim EfficientNet B7 \cite{dfdc_solution}\tnote{\textdagger} & 0.972 & 90.6\% & 462M\\
    Convolutional ViT \cite{wodajo2021deepfake} & 0.843 & 77.0\% & 89M\\
    \midrule
    Efficient ViT (our)     & 0.919  & 83.8\%  & 109M\\
    Conv. Cross ViT Wodajo CNN (our) & 0.925 & 84.5\% & 142M\\
    Conv. Cross ViT Eff.Net B0 - Avg (our) & 0.947 & 85.6\% & 101M\\
    Conv. Cross ViT Eff.Net B0 - Voting (our)     & 0.951 & 88.0\% & 101M\\
    \bottomrule
\end{tabular}
\begin{tablenotes}
    \item[\textdagger] Uses an ensemble of 6 networks.
\end{tablenotes}
\end{threeparttable}
% \addtocounter{footnote}{-2}
\end{table}

\begin{table}[t]
\centering
\caption{Models accuracy on FaceForensics++} 
\label{tab:ff_results}
\resizebox{\textwidth}{!}{
\begin{tabular}{lcccccccc} 
    \toprule
    \textbf{Model}  & Mean & FaceSwap & DeepFakes & FaceShifter & NeuralTextures \\
    \midrule
    Convolutional ViT \cite{wodajo2021deepfake}    & 67\% & 69\%  & \textbf{93\%} & 46\% &60\% \\
    Efficient ViT (our)      & 76\%  & 78\% & 83\% & 76\% & 68\% \\
    Conv. Cross ViT Wodajo CNN (our) & 76\% & 81\% & 83\% & 73\% & 67\% \\
    Conv. Cross ViT EfficientNet B0 (our)     & \textbf{80\%}   & \textbf{84\%} & 87\% & \textbf{80\%} & \textbf{69\%} \\
    \bottomrule
\end{tabular}

}
\end{table}

The models with Cross Vision Transformer show a particularly marked improvement when using the EfficientNet B0 as a patch extractor. Although the AUC and F1-score remain slightly below other state-of-the-art methods (in the first two rows of Table \ref{tab:dfdc_results}), these results were obtained using neither distillation nor ensemble techniques that complicate both training and inference. In fact, we can notice how the Cross Vision Transformer with the EfficientNet extractor can reach a competitive performance using less than 1/3 of the parameters of the top methods. 

\begin{figure}[t]
    \centering
    \includegraphics[width=0.7\textwidth]{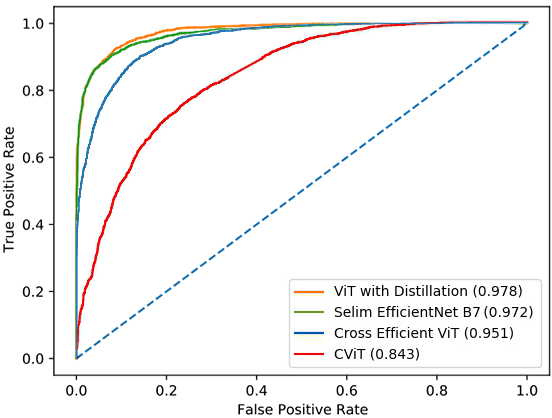}
    \caption{ROC Curves comparison between our best model and others on DFDC test set.}
    \label{figure:roc}
\end{figure}

Furthermore, in the last two rows of Table \ref{tab:dfdc_results} we can notice how our voting procedure used at inference time can slightly improve the results with respect to a plain average of the scores from all the faces indistinctly, as done by the other methods.
In Fig. \ref{figure:roc} we report a detailed ROC plot for the architectures on the DFDC dataset.

In order to compare the developed models also on another dataset, we carried out some tests also on FaceForensics++. %investigating the ability of the obtained networks to correctly identify video deepfakes also on further different deepfakes generation techniques contained in this dataset.
As shown in Table \ref{tab:ff_results}, our models outperform the original Convolutional ViT \cite{wodajo2021deepfake} on all sub-datasets of FaceForensics++, excluding DeepFakes. This is probably because the network could better generalize on very specific types of deepfakes. %but learned to recognise less that specific type of Deepfake. 
It is worth noting how the results obtained in terms of accuracy on the various sub-datasets confirm the assumption already made in \cite{wodajo2021deepfake}: some deepfakes techniques such as NeuralTextures produce videos that are more difficult to find, thus resulting in lower accuracy values than other sub-datasets. % such as Deepfakes. 
However, the average of all our three models is higher than the average obtained by the Convolutional ViT. The Convolutional Cross ViT achieves the best result with the EfficientNet B0 backbone, obtaining a mean accuracy of 80\%. %as the average of the test sets.

\section{Conclusions}
In this research, we demonstrated the effectiveness of mixed convolutional-transformer networks in the Deepfake detection task. Specifically, we used pre-trained convolutional networks, such as the widely used EfficientNet B0, to extract visual features, and we relied on Vision Transformers to obtain an informative global description for the downstream task. We showed that it is possible to obtain state-of-the-art results without necessarily resorting to distillation techniques from models based on convolutional or ensemble networks.
The use of a patch extractor based on EfficientNet proved to be particularly effective even by simply using the smallest network in this category. EfficientNet also led to better results than the generic convolutional network trained from scratch used in Wodajo et al \cite{wodajo2021deepfake}.
We then proposed a mixed architecture, the Convolutional Cross ViT, that works at two different scales to capture local and global details.
The tests carried out with these models demonstrated the importance of multi-scale analysis for determining the manipulation of an image. %This can be a starting point for the realization of future architectures able to exploit these features to obtain even better results.

We also paid particular attention to the inference phase. In particular, we presented a simple yet effective voting scheme for explicitly dealing with multiple faces in a video. The scores from multiple actor faces are first averaged over time, and only then hard voting is used to decide if at least one face was manipulated. This inference mechanism yielded slightly better and stable results than the global average pooling of the scores performed by previous methods. % especially when only one of the many faces in a video is manipulated. %pointed out that various precautions in the testing phase can be crucial for obtaining qualitatively valid and stable results; therefore, 
%we recall that it is required not only to come up with better performing models, but also to create inference strategies that are more optimized and are harder to fool. % that are more difficult to deceive and as optimised as possible.

%\section{Acknowledgements}
%This work was partially supported by “Intelligenza Artificiale per il Monitoraggio Visuale dei Siti Culturali" (AI4CHSites) CNR4C program, CUP B15J19001040004, by the European Commission projects AI4EU (GA n. 825619), and AI4Media (GA n. 951911).

\bibliographystyle{splncs04}
\bibliography{references}  %%% Uncomment this line and comment out the ``thebibliography'' section below to use the external .bib file (using bibtex) .

%%% Uncomment this section and comment out the \bibliography{references} line above to use inline references.
% \begin{thebibliography}{1}

% 	\bibitem{kour2014real}
% 	George Kour and Raid Saabne.
% 	\newblock Real-time segmentation of on-line handwritten arabic script.
% 	\newblock In {\em Frontiers in Handwriting Recognition (ICFHR), 2014 14th
% 			International Conference on}, pages 417--422. IEEE, 2014.

% 	\bibitem{kour2014fast}
% 	George Kour and Raid Saabne.
% 	\newblock Fast classification of handwritten on-line arabic characters.
% 	\newblock In {\em Soft Computing and Pattern Recognition (SoCPaR), 2014 6th
% 			International Conference of}, pages 312--318. IEEE, 2014.

% 	\bibitem{hadash2018estimate}
% 	Guy Hadash, Einat Kermany, Boaz Carmeli, Ofer Lavi, George Kour, and Alon
% 	Jacovi.
% 	\newblock Estimate and replace: A novel approach to integrating deep neural
% 	networks with existing applications.
% 	\newblock {\em arXiv preprint arXiv:1804.09028}, 2018.

% \end{thebibliography}

\end{document}